\documentclass{article}
\usepackage{spconf,amsthm,amsmath,amssymb,graphicx}
\usepackage{comment}
\usepackage{algorithm2e}
\usepackage{algpseudocode}
\usepackage{url}            

\makeatletter

\newcommand{\argmax}{\mathop{\mathrm{argmax}}}
\newcommand{\argmin}{\mathop{\mathrm{argmin}}}
\newcommand{\am}{\text{am}}
\newcommand{\lm}{\text{lm}}
\newcommand{\cost}{\text{cost}}
\newcommand{\ccost}{\text{ccost}}
\newcommand{\next}{\text{next}}
\newcommand{\arcs}{\text{Arcs}}
\newcommand{\pred}{\text{Pred}}
\newcommand{\word}{\text{word}}

\newcommand{\toprule}{\hrule height.8pt depth0pt \kern2pt} 
\newcommand{\midrule}{\kern2pt\hrule\kern2pt} 
\newcommand{\bottomrule}{\kern2pt\hrule\relax}
\newcommand{\algcaption}[2][]{%
	\refstepcounter{algorithm}%
	\@ifmtarg{#1}
	{\addcontentsline{loa}{figure}{\protect\numberline{\thealgorithm}{\ignorespaces #2}}}
	{\addcontentsline{loa}{figure}{\protect\numberline{\thealgorithm}{\ignorespaces #1}}}%
	\toprule
	\textbf{\fname@algorithm~\thealgorithm}\ #2\par 
	\midrule
}
\def\name#1{\gdef\@name{#1\\}}
\makeatother

\title{Lattice Rescoring Strategies for Long Short Term Memory \\ Language Models in Speech Recognition}
\name{{\em Shankar Kumar, Michael Nirschl, Daniel Holtmann-Rice, Hank Liao, Ananda Theertha Suresh, Felix Yu}}
\address{Google Inc., USA}
\begin{document}
%
\maketitle
\begin{abstract}
	Recurrent neural network (RNN) language models (LMs) and Long Short Term Memory (LSTM) LMs, a variant of RNN LMs, have been shown to outperform traditional N-gram LMs on speech recognition tasks. However, these models are computationally more expensive than N-gram LMs for decoding, and thus, challenging to integrate into speech recognizers. Recent research has proposed the use of lattice-rescoring algorithms using RNNLMs and LSTMLMs as an efficient strategy to integrate these models into a speech recognition system. In this paper, we evaluate existing lattice rescoring algorithms along with new variants on a YouTube speech recognition task. Lattice rescoring using LSTMLMs reduces the word error rate (WER) for this task by 8\% relative to the WER obtained using an N-gram LM.
\end{abstract}

\section{Introduction}
\label{sec:intro}A language model (LM) is a crucial component of a statistical speech recognition system~\cite{jelinek1997}. The LM assigns a probability to a sequence of words, ${w}_{1}^{T}$:
\begin{equation}
P({w}_{1}^{T}) = \prod_{i=1}^{T} P(w_i | w_1, w_2, \ldots, w_{i-1}).
\end{equation}
N-gram LMs have traditionally been the language model of choice in speech recognition systems because they are efficient to train and use in decoding. An N-gram LM makes the assumption that the $N^{\text{th}}$ word depends only on the previous $N-1$ words~\cite{jelinek1997}:
\begin{equation}
P({w}_{1}^{T}) = \prod_{i=1}^{T} P(w_i | w_{i-(N-1)}, w_{i-(N-2)}, \ldots, w_{i-1}).
\end{equation}
The efficiency of N-gram LMs stems from the small N-gram orders ($N\le5$) typically used to estimate these models. While this makes the N-gram LMs powerful for tasks such as voice-search where short-range contexts suffice, they do not perform as well at tasks such as transcription of long form speech content, that require modeling of long-range contexts~\cite{mikolov2010}. If the order of the N-gram model is increased to say, $N \ge 8$, the resulting LM is both large in size and poorly trained due to the sparsity of higher order N-gram counts. Neural network language models (NNLM) have recently emerged as a promising alternative to N-gram LMs~\cite{bengio2003,schwenk2007,mikolov2010,arisoy2012,sundermeyer2015,liu2016,josefowicz2016,damavandi2016}. By embedding words in a continuous space, an NNLM provides a better smoothing for unknown as well as rare words relative to an N-gram LM~\cite{josefowicz2016}. In addition, a recurrent NNLM (RNNLM) has the ability to model long-range contexts by effectively making use of the entire word history, which results in improved transcription of long form speech content~\cite{mikolov2010,mikolov2011b}. A Long Short Term Memory LM (LSTMLM) ~\cite{hochreiter1997,sundermeyer2015,josefowicz2016} is a variant of RNNLM that does not suffer from the problem of vanishing or exploding gradients~\cite{bengio1994} and has been shown to be more effective for speech recognition than a basic RNNLM~\cite{sundermeyer2015}.

\begin{figure}[hb]
	\centering
	\includegraphics[height=1.5in]{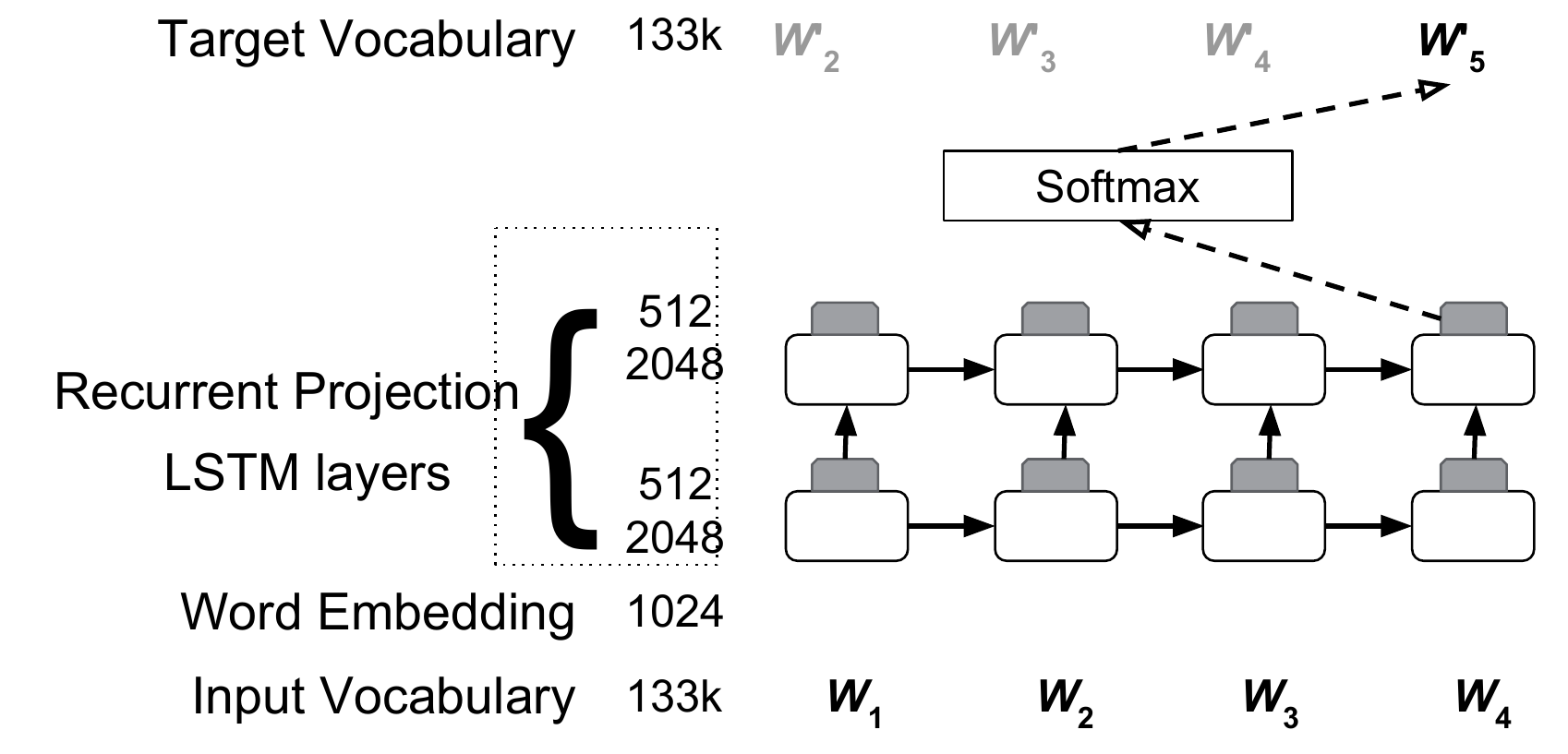}
	\caption{Architecture of the LSTMLM: State of the network while predicting the final target word $w_{5}$, is shown.}
	\label{fig:lstmlm}
\end{figure}

Because LSTMLMs make use of long range word history for predicting word probabilities, they are challenging to integrate into the first pass of a speech recognition system~\cite{sundermeyer2015}. Typically, a $K$-best list of recognition hypotheses is generated in the initial recognition pass and then rescored using the LSTMLM. However for transcribing long form content, a $K$-best list, where $K\le10,000$, represents only a tiny subset of the possible hypotheses. An alternative is to perform rescoring of lattices, which contain many more hypotheses relative to $K$-best lists. This is computationally prohibitive because an LSTMLM uses the full word history for computing the probability of the next word. Recent work~\cite{sundermeyer2015,liu2016} has demonstrated efficient strategies for lattice rescoring. We extend this work and compare both existing algorithms and new variants to perform lattice rescoring with LSTMLMs for transcribing videos from YouTube. We also present techniques for speeding up training and inference of LSTMLMs and for reducing the memory requirements at inference.
\section{Model Architecture}

We adopt the LSTMLM architecture from ~\cite{josefowicz2016,sak2014} (Figure~\ref{fig:lstmlm}). Each cell within the LSTM layers in Figure~\ref{fig:lstmlm} is specified as:
\begin{align}
& & \\ 
\nonumber i_t & = & \sigma(W_{xi} x_t + W_{ri} r_{t-1} + D_{wi} c_{t-1} + b_i) \\
\nonumber f_t & =  & 1.0 - i_t \\
\nonumber o_t & = & \sigma(W_{xo} x_t + W_{ro} r_{t-1} + D_{wo} c_{t-1} + b_o) \\
\nonumber c_t & = & c_{t-1} \odot f_t   + i_t \odot \text{tanh}(W_{xc} x_t + W_{rc} r_{t-1} + b_c) \\
\nonumber m_t & = & \text{tanh}(c_t) \odot o_t \\
\nonumber r_t & = & W_{rm} m_t 
\end{align}
where $x_t = E w_t$ is a continuous representation of the word $w_t$ obtained using the embedding matrix $E$. $\sigma$ is the logistic sigmoid function. $i_t$, $f_t$ and $o_t$ are the \textit{input}, \textit{forget} and the \textit{output} gates. The \textit{forget} and \textit{input} gates are coupled to limit the range of the internal state of the cell. 
The tuple $(c_t, r_t)$ represents the internal cell states or the \textit{LSTM state}. $W$, $D$ and $b$ denote full weight matrices, diagonal weight matrices and biases, respectively. $\odot$ represents element-wise multiplication. The architecture uses \textit{peep-hole} connections from the internal cells ($c_t$) to the gates to learn the precise timings of the outputs~\cite{graves2013,gers2003}. We use a recurrent projection layer ($r_t$)~\cite{sak2014} that reduces the dimension of a big recurrent state ($m_t$), thus allowing the matrices to remain relatively small while retaining a large memory capacity. 
\section{Training}
\label{sec:train}
A large vocabulary speech recognition task typically employs a vocabulary of several hundreds of thousands of words. For such a large vocabulary, the softmax layer is computationally expensive at both training and test times. 

To reduce computation in training, we approximate the softmax over the full vocabulary with a \textit{sampled softmax} over 3-6\% of the words in the vocabulary~\cite{josefowicz2016,abadi2016}. After sorting the words in descending order of their frequencies in the training corpus, we draw the negative samples (i.e. words except for the predicted word) from a log-uniform distribution. Under a \textit{sampled softmax}, the probability of a word $w_i$ given the history $h_i = [w_1, w_2, \ldots, w_{i-1}]$ of previous words is given by:
\begin{align}
\label{eq:softmax} P(w_i|h_i) & = & \frac{e^{U(w_i,h_i)}}{\underset{w' \in V}{\sum} e^{U(w',h_i)}} \propto & \frac{e^{U(w_i,h_i)}}{\underset{w' \in S \cup \{w_i\}}{\sum} e^{U(w',h_i)}},
\end{align}
where $U(w,h)$ is the output of the final LSTM layer (Figure~\ref{fig:lstmlm}) and $S$ is a subset of the vocabulary $V$, excluding the predicted word $w_i$, drawn randomly for each mini-batch of training examples.

At inference time, most of the computation in an LSTMLM takes place in the softmax layer. Multiple schemes have been proposed to reduce the computation in the softmax layer~\cite{morin2005,mikolov2011a,shi2014,chen2016}. We use the self-normalization strategy~\cite{devlin2014} whereby the objective function for training the LSTMLM is modified by adding a penalty if the normalizers $Z(h_i) =  \sum_{w' \in V} e^{U(w',h_i)}$  are different from 1:
\begin{equation}
\sum_{i=1}^{T} \text{log} P(w_i|h_i) + \alpha \: {\text{log}}^2 \left(\sum_{w' \in V} e^{U(w',h_i)}\right),
\end{equation}
where $(w_1, h_1), \ldots, (w_T, h_T)$ is the training data. Self-normalization yields normalizers $Z(h)$ that approximate 1.0. This enables us to obtain large savings in computation by skipping the softmax computation (Equation~\ref{eq:softmax}) at inference time. However, training the model using this scheme is slow because computing the normalizer involves a summation over words in the full vocabulary. To speed up training, we compute the normalizer only over the randomly drawn sample of words within each training mini-batch:
\begin{equation}
\sum_{i=1}^{T} \left\{\text{log} P(w_i|h_i) + \alpha \: {\text{log}}^2 \left(\sum_{w' \in S \cup \{w_i\}} e^{U(w',h_i)}\right)\right\},
\label{eq:selfnorm}
\end{equation}
where the optimal value of $\alpha$ is chosen via a grid search. Values in the range $(0.005-0.01)$ work well in practice because they provide a good trade-off between achieving self-normalization and staying close to the original likelihood objective function.
\section{Decoding}
\label{sec:decode}
Given an acoustic observation sequence $O$, a speech recognizer selects the hypothesis with the highest posterior probability:
\begin{equation}
  \hat{W} = \argmax_{W} P(W|O) = \argmax_{W} P(O|W) P(W),
\end{equation}
where $P(O|W)$ and $P(W)$ are the acoustic and the language model probabilities respectively. To reduce the computation involved in integrating an LSTMLM into the first pass of a recognizer~\cite{sundermeyer2015,liu2016}, we generate lattices using an initial recognition pass that incorporates the acoustic model and the N-gram LM. In this paper, our goal was to evaluate LSTMLMs for speech recognition of long form content such as videos, where the average segment length was 94 words. We focus on lattice rescoring with LSTMLMs that allows us to search over many more hypotheses than a K-best list but requires an approximation.

$Q_L$ denotes the set of nodes in a lattice $L$. $\pred(q, L)$ is the set of nodes that immediately precede $q$ in $L$. $\arcs(q, L)$ is the set of arcs starting from a node $q$ in $L$. For an arc $a$, $a.\am$, $a.\lm$, $a.\word$ and $a.\next$ refer to the acoustic model costs, language model costs, word on the arc and the destination node of the arc, respectively. An arc from node $q'$ to $q$ is denoted as $(q',q)$. 
\begin{figure}[ht]
	\centering
	\includegraphics[height=1.8in]{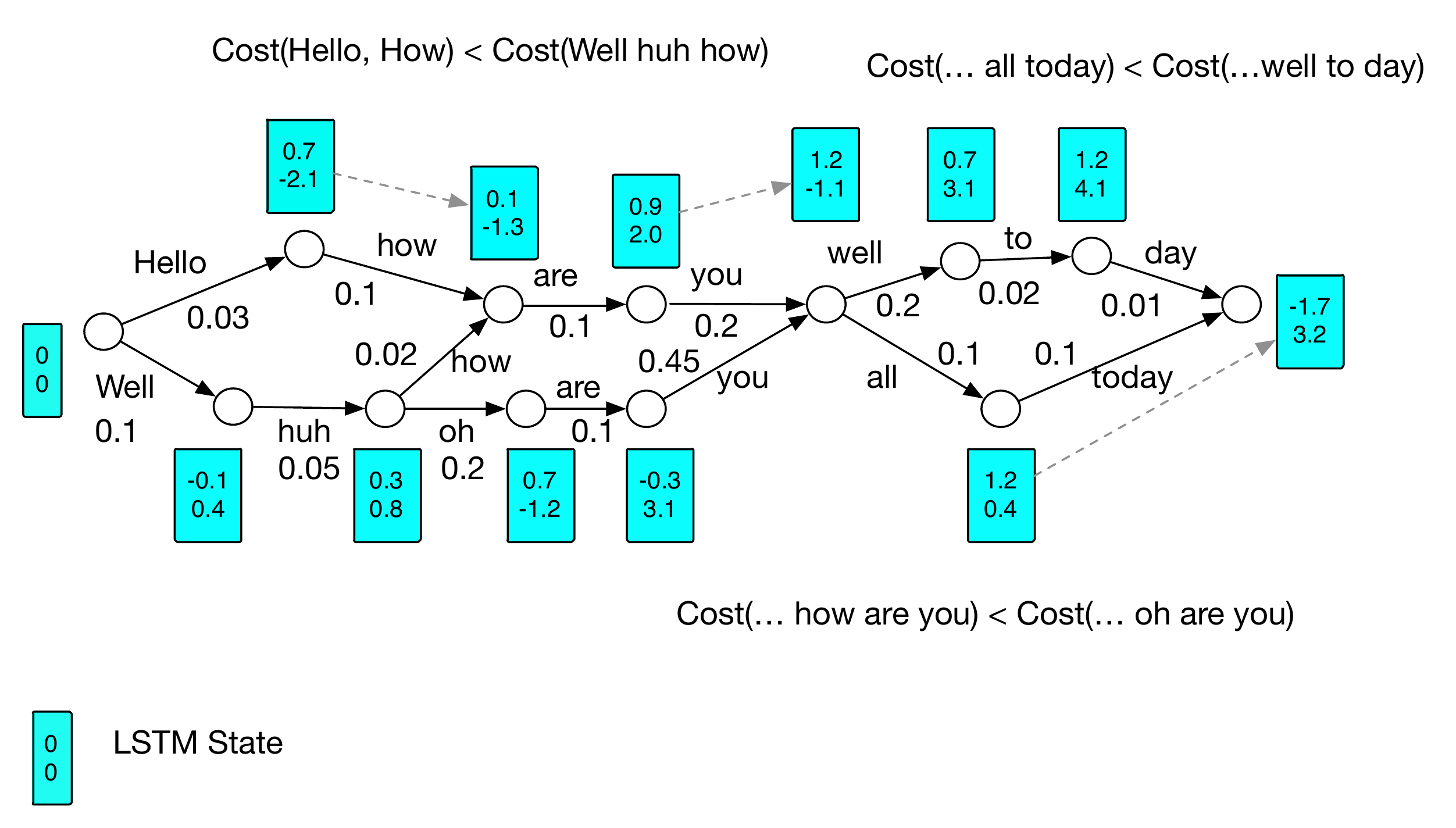}
	\caption{\textit{Push-forward} algorithm with $k=1$ for a toy lattice. The LSTM state is shown as a 2-dimensional vector. At each lattice node, the LSTM state with the lowest cumulative cost is propagated to the outgoing arcs. }   
	\label{fig:pushforward}
\end{figure}
\subsection{Push-forward algorithm}
Algorithm~\ref{algo:pushforward} describes the \textit{push-forward} algorithm~\cite{auli2011} adapted to lattices generated from a speech recognition system. The algorithm extends the LSTM state by the words on lattice arcs while retaining the top $k$ LSTM states at each lattice node. The LSTM states are sorted based on the cumulative acoustic and language model probabilities leading up to the lattice node. The algorithm constructs a new lattice which has the LSTMLM costs on its arcs. For the special case of $k=1$, the original lattice structure can be preserved, leading to a very efficient algorithm (terFigure~\ref{fig:pushforward}). However in doing so, we have approximated the set of LSTM states leading to a lattice node with the single-best LSTM state. We can make this approximation less severe by expanding the initial lattice to a specified N-gram order
prior to applying the \textit{push-forward} algorithm. This expansion ensures that the paths leading to a lattice node end with the same $N-1$ words. This approach is similar to the recombination pruning in~\cite{sundermeyer2015} and the N-gram history based clustering approach of ~\cite{liu2016}. 
\subsection{LSTM state pooling algorithm}
We next study a variant of the rescoring algorithm proposed in~\cite{ladhak2016} whereby the original structure of the lattice is preserved but instead of keeping the most likely LSTM state at each lattice node, the LSTM state at a given lattice node is computed as a weighted combination of the LSTM states of its predecessor nodes.
Our variant differs from~\cite{ladhak2016} in the use of alternative weighting strategies. The pooling algorithm makes use of the continuous representation of the word histories, encoded by the LSTM state:

\begin{equation}
h_{\text{LSTM}}(q) = \sum_{q' \in \pred(q, L)} h_{\text{LSTM}}(q') P(q|q').
\end{equation}
The probability of transitioning from $q'$ to $q$ is estimated from the weight of lattice paths ending at $q'$: 
\begin{equation}
P(q|q') = \frac{\text{Weight}(q')}{\sum_{q'' \in \pred(q, L)} \text{Weight}(q'')},
\end{equation}
where $\text{Weight}(q')$ is defined as either the highest probability over all lattice paths $W_{q'}$ terminating at $q'$:
\begin{equation}
\label{eq:poolmaxprob}
\text{Weight}(q') = \max_{W_{q'}} P(O|W_{q'}) P(W_{q'}),
\end{equation}
or the sum of probabilities of all lattice paths ending at $q'$:
\begin{equation}
\label{eq:poolsumprob}
\text{Weight}(q') = \sum_{W_{q'}} P(O|{W}_{q'}) P(W_{q'}).
\end{equation}
Both weights (Eqns~\ref{eq:poolmaxprob} and ~\ref{eq:poolsumprob}) can be computed efficiently using a forward-backward algorithm~\cite{wessel2001}. We compare these two weighting schemes with a simpler scheme that assigns a uniform probability to each of the predecessor nodes. i.e.
\begin{equation}
P(q|q') = \frac{1}{|\pred(q, L)|} \forall q' \in \pred(q, L).
\label{eq:pooluniformprob}
\end{equation}
Analogous to the \textit{push-forward} algorithm with $k=1$, the \textit{LSTM state pooling} algorithm also preserves the structure of the input lattice, thus allowing for efficient computation (Algorithm~\ref{algo:statepooling}).
\subsection{Arc Beam algorithm}
The \textit{push-forward} algorithm with $k=1$ selects the best LSTM state at each lattice node and uses that LSTM state for scoring the outgoing arcs from that node. We propose an \textit{arc-beam} variant that allows each arc to independently select the best LSTM state from among the predecessor nodes, and use this state for scoring the word on the arc. In this algorithm, it is possible that two arcs starting at the same lattice node can select different LSTM states for scoring (Algorithm~\ref{algo:arcbeam}).
  \begin{algorithm}
	\SetAlgoLined
	\KwData{$k, L, LSTMLM$}
	\KwResult{$\hat{W}$}
	$Q \leftarrow \text{Topological Sort}(Q_{L})$\\
	$h_0.s = 0 $, 
	$h_0.w = \epsilon$, 
	$h_0.\am = h_0.\lm = h_o.\cost = 0$ \\ \Comment{Initialize LSTM state to 0, previous word to empty ($\epsilon$), AM/LM/total costs to 0}\\
	$H_s.ADD(h_0)$ \Comment{Add initial hypothesis to start node} \\
	\For{$q \in Q$} {
		$H_q \leftarrow \text{Top}(k, H_q)$ \Comment{Top $k$ states by $h.\cost$} \\
		\For{$h \in H_q$}{
			\For{$a \in \arcs(q, L)$}{
				\begin{flushleft}
					$(s', \bf{p}) \leftarrow \text{LSTMLM}(h.s, h.w)$ \Comment{Inference}
					$h' = \{\}$  \Comment{Create new hypothesis}\\
					$h'.s = s'$ \Comment{LSTM state} \\
					$h'.\am =  a.\am$ \\
					$h'.\lm =  -log(\bf{p}[a.word])$ \\
					$h'.\cost = h.\cost + a.\am + a.\lm$\\\Comment{Total cost}\\
					$h'.w = a.word$\\
					$\text{parent}(h') = h$\Comment{Keep backpointers}\\
					$H_{a.\next}.ADD(h')$
				\end{flushleft}
			}
		}
	}
	Create a rescored lattice $R$ based on backpointers in $H$ \\
	$\hat{W} = \text{Best Cost}(R)$\\\Comment{Best path in $R$ using costs in $a.\cost$}
   \caption{\textit{Push-forward} algorithm~\cite{auli2011}.}
	\label{algo:pushforward}
  \end{algorithm}
 \begin{algorithm}
	\SetAlgoLined
	\KwData{$L, LSTMLM$}
	\KwResult{$\hat{W}$}
	$Q \leftarrow \text{Topological Sort}(Q_{L})$\\
	$h[0] = 0$ \Comment{Initialize LSTM state} \\
	$\forall q \in Q, \cost[q] = \infty$ \Comment{Initialize costs} \\
	$\cost[0] = 0$ \Comment{Cost at start node} \\
	$w[0] = \epsilon$ \Comment{Previous word at start node} \\
	\For{$q \in Q$} { 
		\For{$a \in \arcs(q, L)$}{
			$(s', \bf{p}) \leftarrow \text{LSTMLM}(h[q], w[q])$
			$a.\lm = -log(\bf{p}[a.word])$
			$\ccost = \cost[q] + a.\am + a.\lm$ \Comment{Total cost} \\
			\If{$\cost[a.\next] > \ccost$} { \Comment{Keep best previous word at each state} \\
				$\cost[a.\next] = \ccost$ \\
				$w[a.\next] = a.\word$
			}
			$h[a.\next] += s' * \text{weight}(q,a.\next)$ \\ \Comment{Pool LSTM states}
		}
	}
	$\hat{W} = \text{Best Cost}(L)$\\\Comment{Best path in $L$ using $\cost[q], q \in Q$}
    \caption{\textit{LSTM State pooling} algorithm~\cite{ladhak2016}}
	\label{algo:statepooling}
\end{algorithm}
\begin{algorithm}[h]
	\SetAlgoLined
	\KwData{$L, LSTMLM$}
	\KwResult{$\hat{W}$}
	$Q \leftarrow \text{Topological Sort}(Q_{L})$\\
	Add a dummy initial node $q_0$ \\
	$h[q_0, 0] = 0$ \\
	$\cost[q_0,0] = 0$\\\Comment{Initialize LSTM state, cost of dummy edge: $(q_0,0)$} \\
	\For{$q \in Q$} { 
		\For{$a \in \arcs(q, L)$}{ 
			$\lm \leftarrow \{\}$ \\
			$ccost \leftarrow \{\}$ \\
			\For{$q' \in \pred(q, L)$}{
				\begin{flushleft}
				$(s', \bf{p}) \leftarrow \text{LSTMLM}(h[q',q], (q',q).\word)$
				$\lm[q'] = -log(\bf{p}[a.word])$
                $\ccost[q'] = \cost[q',q] + a.\am + \lm[q']$ \\ \Comment{Total cost}
                \end{flushleft}
			}
			\begin{flushleft}
				$\hat{q} = \argmin_{q'} \ccost[q']$ \Comment{best previous state} \\
				$\hat{h} = h[\hat{q}, q]$ \Comment{best previous LSTM state}
			  $a.\lm =  \lm[\hat{q}]$ \Comment{Set arc cost}
			  $h[q, a.nextstate] = \hat{h}$ \\ \Comment{Best LSTM state for arc}
                          $\cost[q, a.\next] = \ccost[\hat{q}]$ \\
			\end{flushleft}
	 }
	}
	$\hat{W} = \text{Best Cost}(L)$\Comment{Best path in $L$ using $\cost[q,q']$}
        \caption{\textit{Arc Beam} algorithm.}
	\label{algo:arcbeam}
\end{algorithm}
\subsection{Handling Non-Speech Symbols}
Our lattices contain arcs with non-speech symbols representing pauses or sentence boundaries. However, the LSTMLM is not trained on such tokens. To prevent a train-test mismatch, we modify the lattice rescoring algorithms to copy the LSTM state across such tokens. Such a strategy is also employed in~\cite{ladhak2016}.

\section{Model Compression}
For a speech recognition task with a vocabulary $>100k$ words, it is crucial to train an LSTMLM with large embedding and softmax layers to achieve good performance. With a vocabulary size of 133k words and an embedding dimension of 1024 nodes, there are $133,000 \times 1024 = 136M$ parameters. Similarly, if the input to the softmax has 512 dimensions, then the softmax layer will have $133,000 \times 512 = 68M$ parameters. In our LSTMLM (Figure~\ref{fig:lstmlm}), the parameters in the model that are not part of either the embedding or the softmax layers, do not contribute substantially to the overall size of the model. To compress the LSTMLM, we focused on shrinking the embedding and the softmax matrices.

Scalar quantization has been shown to be effective in compression neural network models~\cite{brainmtwhitepaper}. However, the resulting compression ratios were not adequate for our task. An alternative strategy is vector quantization (VQ), which can lead to larger compression. However, VQ does not work well in high dimensional settings consisting of say, 1024 dimensions. \textit{Product Quantization} (PQ)~\cite{sabin1984} is a hybrid strategy that allows us to apply VQ in higher dimensions. In prior work~\cite{gong2014,wu2016}, PQ has been applied to compress the fully connected and convolutional layers of neural networks. In contrast, we apply PQ to compress the embedding and softmax layers of the LSTMLM. Others~\cite{joulin2016} have also applied PQ to compress word embedding layers for text classification models.
For applying PQ to the embedding (softmax) matrix, we divide the embedding (softmax) dimensions into equal sized chunks and perform VQ on each chunk. As a result, we need to store only the VQ indices and the centers of each VQ codebook. Using a chunk size of either 4 or 8, we were able to shrink the embedding and softmax matrices by a factor of 16 or 32 without any loss in performance of the LSTMLM. With PQ, the matrix-matrix multiplication in the softmax layer can be efficiently performed using a lookup table. PQ is performed subsequent to model training.
\section{Experiments}
We conducted our experiments on the YouTube speech recognition task~\cite{liao2013,soltau2017}. YouTube is a video sharing website with over a billion users. Speech recognition has enabled YouTube to improve accessibility to users by providing automatic captions. Our language models were trained on 4.8 billion word tokens derived from anonymized user-generated video captions. Our evaluation set consisted of 296 videos sampled randomly from the \textit{Google preferred channels} on YouTube~\cite{googlepreferred}.  It contained 13 categories, ranging from Science\&Education to Video Games~\cite{liao2013, kuznetsov2016, soltau2017}. It had a total duration of 25 hours and 250,000 word tokens.

We divided the audio portion of each video into segments using a Gaussian mixture model based speech-silence endpointer. For each segment, we generated an initial word lattice using a context dependent phone level acoustic model trained using the connectionist temporal classification (CTC) objective function as described in~\cite{soltau2017} and an N-gram LM with a recognition vocabulary of 947K words. We then rescored the lattice using the LSTMLM. To investigate the effect of the order of the N-gram LM used for lattice generation on the performance of the LSTMLM lattice rescoring algorithms, we experimented with a bigram LM and a 5-gram LM, each consisting of 23M N-grams. We used a subset of the first pass vocabulary, consisting of the most frequent 133,000 words, for training the LSTMLM. The LSTMLM parameters are shown in Figure~\ref{fig:lstmlm}. In training, words in the training set, not present in the 133k vocabulary, were mapped to a special UNK token. During decoding, if a word in the lattice was not present in the 133k vocabulary, it was assigned a probability $P_{\text{LSTM}}(UNK|history) \cdot \frac{1.0}{N_{\text{UNK}}}$, where $N_{\text{UNK}}$ is the number of unique words that were mapped to UNK in training. 

Using hyperparameters from~\cite{josefowicz2016}, we trained the model until convergence using an AdaGrad optimizer~\cite{duchi2011} using a learning rate of 0.2 without \textit{dropout}~\cite{srivastava2013}. We unrolled the RNNs for 20 steps, used a batch size of 128, and clipped the gradients of the LSTM weights so that their norm was bounded above by 1.0~\cite{pascanu2013}. The training used 32 GPU workers and asynchronous gradient updates. Training was performed using TensorFlow~\cite{abadi2016}. For comparison, we also trained a larger 5-gram LM consisting of 1.1 billion N-grams with the same 133k vocabulary used in training the LSTMLM.
\subsection{Perplexity}
We first report perplexity of the LMs on a development set consisting of 142k tokens, when all LMs are trained using the same vocabulary of 133k words (Table~\ref{tab:ppl}). The LSTMLM improves upon the perplexity of the N-gram models by more than 50\%. 
\begin{table}[h]
	\begin{center}	
		\begin{tabular}{c|c|c}
			\hline
			& Number of parameters & Perplexity \\ \hline
            2-gram & 23M N-grams & 332 \\
			5-gram & 23M N-grams & 225 \\
			5-gram & 1.1B  N-grams &  178 \\
			LSTMLM & 222M weights & 80 \\ \hline	
		\end{tabular}
		\caption{{\it Perplexity of LSTMLM vs. N-gram LMs using a 133k word vocabulary.}}
		\label{tab:ppl}
	\end{center}
\end{table}
\subsection{Recognition}
We next report recognition results in terms of Word Error Rate (WER) (Table~\ref{tab:wer}) by rescoring lattices generated using either a bigram or a 5-gram LM.
For LSTMLM, we obtain results from (a) rescoring a $10,000$ best list of unique recognition hypotheses generated from the initial lattice and, (b) by applying the \textit{push-forward} lattice rescoring algorithm with $k=1$ (Algorithm~\ref{algo:pushforward}). The probability computed by the LSTMLM was used as-is and not interpolated with the probability from the first pass LM. We used a language model scale factor of 1.0 and a word insertion penalty of 0.0. Lattices were pruned to a density of 20 using forward-backward pruning~\cite{sixtus1999} prior to rescoring. Using K-best rescoring, LSTMLM yielded only a tiny improvement of $<$0.1\% relative. In contrast, with lattice rescoring, LSTMLM gave a 8.1\% relative improvement over the first pass LM. This confirms that the $10,000$ best list is a tiny space of hypotheses for the video segment. As expected, LSTMLM gives a much larger relative improvement over a bigram first pass LM.
\begin{table}[h]
 {\small
	\begin{center}	
		\begin{tabular}{c|c|c|c|c} \hline
                    Model & Num. parameters & Pass & \multicolumn{2}{c}{WER (\%)} \\ \hline
                  & & & \multicolumn{2}{c}{LM used in} \\ 
                  & & & \multicolumn{2}{c}{lattice generation} \\ \hline
                  & & & 2-gram & 5-gram \\ \hline
                  Initial LM & 23M N-grams & Initial & 14.3 & 13.5 \\ 
                  Large 5-gram & 1.1B N-grams & Lattice & 13.0 & 12.9 \\ \hline
                  LSTMLM & 220M weights & 10k list & 14.1 & 13.4 \\
                  LSTMLM & 220M weights & Lattice & 12.8 & 12.4 \\ \hline
		\end{tabular}
		\caption{{\it WER of LSTMLMs vs. N-gram LMs}}
		\label{tab:wer}
	\end{center}
}
\end{table}

We next present results using lattice rescoring algorithms presented in Section~\ref{sec:decode} (Table~\ref{tab:werlatrescore}). We first varied the maximum hypotheses per lattice node ($k$) in the \textit{push-forward} algorithm 
(Algorithm~\ref{algo:pushforward}) by increasing it from 1 to 50. While the WER reduced by 2\% relative for the 2-gram lattices, it showed no variation for the 5-gram lattices.
The \text{push-forward} algorithm with $k=1$ results in a more severe approximation for the 2-gram lattices. Hence, increasing the LSTM hypotheses per lattice node ($k$) results in a better WER. We next expanded the lattice to different N-gram orders, up to a maximum of 6, prior to applying the \textit{push-forward} algorithm with $k=1$. While 2-gram lattices benefitted considerably from lattice expansion (3\% relative), there was less gain for 5-gram lattices which already contained unique 5-gram histories for most lattice nodes prior to expansion. \textit{LSTM state pooling} (Algorithm~\ref{algo:statepooling}) gives a similar performance as \textit{push-forward} algorithm with $k=1$. Both max-prob (Equation~\ref{eq:poolmaxprob}) and the sum-prob (Equation~\ref{eq:poolsumprob}) weighting schemes performed better than 
assigning a uniform probability to each of the predecessor nodes. Finally, the \textit{arc beam} algorithm (Algorithm~\ref{algo:arcbeam}) performed equivalently to \textit{push-forward} algorithm with $k=1$.
\begin{table}[h]
{\small
	\begin{center}	
		\begin{tabular}{c|cc}
			\hline
			Algorithm & WER (\%) \\ \hline
			& \multicolumn{2}{c}{LM used in} \\
			& \multicolumn{2}{c}{Lattice generation} \\ \hline
			& 2-gram & 5-gram \\ \hline
			\multicolumn{2}{c}{\textit{Push-forward} (Algorithm~\ref{algo:pushforward})} \\ \hline
			$k=1$ & 12.8 & 12.4 \\
			$k=10$ & 12.7 & 12.4 \\
			$k=50$ & 12.6 & 12.4 \\ \hline  
			\multicolumn{2}{c}{Expand lattice to N-gram order, \textit{Push-forward}, $k=1$} \\ \hline
			$N\le2$ & 12.8 & 12.4 \\
			$N=$3 & 12.6 & 12.4 \\
			$N\ge$4 & 12.4 & 12.3 \\ \hline
			\multicolumn{2}{c}{\textit{LSTM State Pooling} (Algorithm~\ref{algo:statepooling})} \\ \hline
			Uniform (Equation~\ref{eq:pooluniformprob}) & 13.0 & 12.6 \\
			Max Prob (Equation~\ref{eq:poolmaxprob}) & 12.8 & 12.5 \\
			Sum Prob (Equation~\ref{eq:poolsumprob}) & 12.8 & 12.5 \\ \hline
			\multicolumn{2}{c}{\textit{Arc Beam} (Algorithm~\ref{algo:arcbeam})} \\ \hline
			Default setting & 12.8 & 12.5 \\ \hline
		\end{tabular}
		\caption{{\it WER of LSTMLM lattice rescoring algorithms}}
		\label{tab:werlatrescore}
	\end{center}
}
\end{table}

\subsection{Self Normalization}
The model trained using self-normalization improved the \textit{real-time factor}, defined as the ratio of the decoding time to the duration of the audio, by a factor of 1.7 with a 0.1\% absolute (0.08\% relative) degradation in WER. 

\section{Discussion}
In this paper, we presented a comparison of algorithms for rescoring speech recognition lattices using an LSTM language model. We obtain an 8\% relative improvement in WER over a first pass N-gram LM on speech recognition of YouTube videos. While the \textit{push-forward} algorithm and its variants have been explored previously~\cite{sundermeyer2015,liu2016}, our work is novel in that we compare these algorithms to \textit{LSTM state pooling}~\cite{ladhak2016} that was proposed for spoken language understanding and has not been evaluated in speech recognition. Pooling LSTM states over the predecessor nodes gives a similar performance as choosing the single best LSTM state at each lattice node. This indicates that the terminal LSTM states of paths ending at a lattice node are similar. We propose the \textit{arc-beam} algorithm, a novel variant of the \textit{push-forward} algorithm that selects the best predecessor LSTM state independently for each lattice arc.  All lattice rescoring algorithms substantially improve upon rescoring of 10K recognition hypotheses. 
This shows that it is crucial to explore a large space of hypotheses in rescoring on a task such as YouTube speech recognition where the average length of an utterance is 94 words. Of the prior lattice rescoring approaches, we did not experiment with the history vector based clustering~\cite{liu2016} because it was shown to give equivalent results to N-gram based history clustering, which, in turn, is similar to applying \textit{push-forward} algorithm on a lattice expanded to a specified N-gram order. Our lattices do not have accurate start and end times on the lattice arcs due to decoder optimization, we therefore do not explore lookahead based techniques that make use of the arc-level start and end times~\cite{sundermeyer2015}. 

Our paper also provides two directions for LSTMLM model optimization. We are the earliest to show that an LSTMLM can be trained using self-normalization~\cite{devlin2014} with a sampled softmax loss~\cite{josefowicz2016}. This allows for efficient training as well as inference, both of which are especially critical for language models with very large vocabularies used in commerical speech recognizers~\cite{damavandi2016}. Second, product quantization is an effective strategy that can help compress both embedding and softmax layers in an LSTMLM by a factor of 16 to 32 without any loss in performance. This leads to a large reduction in the overall size of the LSTMLM, in which the embedding and softmax layers contribute to the bulk of the model size. We expect our lattice rescoring and model optimization strategies to be useful for speech recognition and related tasks such as handwriting recognition where long-span language models have proven to be useful.
\section{Acknowledgements}
We thank Jitong Chen, Michael Riley, Brian Roark, Hagen Soltau, David Rybach, Ciprian Chelba, Chris Alberti, Felix Stahlberg and Rafal Jozefowicz for helpful suggestions. 
\newpage

\bibliographystyle{IEEEbib}
\bibliography{lstmlm}
	
\end{document}